\documentclass{article}
\usepackage{graphicx} 
\usepackage{booktabs}
\usepackage{array}
\usepackage{multirow}
\usepackage{colortbl}
\usepackage{makecell}
\usepackage[ruled,linesnumbered]{algorithm2e}
\usepackage[linesnumbered, ruled]{algorithm2e}
\usepackage{authblk}
\usepackage{caption2}

\usepackage{amsmath}

\title{ChatABL: Abductive Learning via Natural Language Interaction with ChatGPT}

\author[1]{Tianyang Zhong}
\author[1]{Yaonai Wei}
\author[1]{Li Yang}
\author[2]{Zihao Wu}
\author[2]{Zhengliang Liu}
\author[1]{Xiaozheng Wei}
\author[1]{Wenjun Li}
\author[8]{Junjie Yao}
\author[1]{Chong Ma}
\author[3]{Xiang Li}
\author[4]{Dajiang Zhu}
\author[8]{Xi Jiang}
\author[1]{Junwei Han}
\author[5,6,7]{Dinggang Shen}
\author[2]{Tianming Liu}
\author[1]{Tuo Zhang \thanks{Corresponding author: tuozhang@nwpu.edu.cn}}

\affil[1]{\normalsize \emph{School of Automation, Northwestern Polytechnical University, Xi'an 710072, China} }
\affil[2]{\emph{School of Computing, The University of Georgia, Athens 30602, USA}}
\affil[3]{\emph{Department of Radiology, Massachusetts General Hospital and Harvard Medical School, Boston 02115, USA}}
\affil[4]{\emph{Department of Computer Science and Engineering, The University of Texas at Arlington, Arlington 76019, USA}}
\affil[5]{\emph{School of Biomedical Engineering, ShanghaiTech University, Shanghai 201210, China}}
\affil[6]{\emph{Shanghai United Imaging Intelligence Co., Ltd., Shanghai 200230, China}}
\affil[7]{\emph{Shanghai Clinical Research and Trial Center, Shanghai, 201210, China}}
\affil[8]{\emph{School of life science and technology, University of Electronic Science and Technology of China, Chengdu 611731, China}}

\date{}

\begin{document}
\begin{sloppypar}
\maketitle

\vspace{-3em}

\abstract{Large language models (LLMs) such as ChatGPT have recently demonstrated significant potential in mathematical abilities, providing valuable reasoning paradigm consistent with human natural language. However, LLMs currently have difficulty in bridging perception, language understanding and reasoning capabilities due to incompatibility of the underlying information flow among them, making it challenging to accomplish tasks autonomously. On the other hand, abductive learning (ABL) frameworks for integrating the two abilities of perception and reasoning has seen significant success in inverse decipherment of incomplete facts, but it is limited by the lack of semantic understanding of logical reasoning rules and the dependence on complicated domain knowledge representation. This paper presents a novel method (ChatABL) for integrating LLMs into the ABL framework, aiming at unifying the three abilities in a more user-friendly and understandable manner. The proposed method uses the strengths of LLMs’ understanding and logical reasoning to correct the incomplete logical facts for optimizing the performance of perceptual module, by summarizing and reorganizing reasoning rules represented in natural language format. Similarly, perceptual module provides necessary reasoning examples for LLMs in natural language format. The variable-length handwritten equation deciphering task, an abstract expression of the Mayan calendar decoding, is used as a testbed to demonstrate that ChatABL has reasoning ability beyond most existing state-of-the-art methods, which has been well supported by comparative studies. To our best knowledge, the proposed ChatABL is the first attempt to explore a new pattern for further approaching human-level cognitive ability via natural language interaction with ChatGPT.}

\section{Introduction}\label{introduction}

Large Language Models (LLMs), which learn from unprecedented levels of data to generate human-like responses, have emerged as advanced artificial general intelligence (AGI) systems  \cite{blum2023theoretical,brown2020language,li2023wisdom,liu2023deid,wang2023chatcad}. LLMs play a pivotal role in language translation  \cite{peng2023towards}, problem-solving  \cite{liu2023summary}, naming entity recognition \cite{rezayi2022clinicalradiobert}, and text generation \cite{dai2023chataug}. Inspired by their remarkable progress in natural language processing, it will be an interesting topic to bridge the capacities of perception, language understanding and reasoning (PLR) to explore advanced intelligent behaviors of humans, which has profound significance for the development of new AGI systems.


From the perspective of human perception and cognitive ability \cite{chan1997reactions,humphrey1985work,nes2023perception,williams2023hierarchical,zhu2010heritability}, image processing includes understanding the spatial relationship between objects, identifying patterns and textures, and extracting features that describe objects in images \cite{biederman1985human,wang2023chatcad}. More importantly, these results are further executed through rigorous logical reasoning \cite{ganapini2022combining,sloman1996empirical,xu2023mind}, and language interaction to achieve information exchange with the external world. These tasks require a deep collaborative understanding of perception, language understanding and reasoning components, which is challenging for LLMs that have been primarily trained on text data \cite{bubeck2023sparks}. This limitation is mainly manifested in how to establish an effective communication framework to satisfy the compatibility of underlying information flow among them, which provides inspiration for further research, that is, more attention should be paid to the structural design of the model on the basis of LLMs \cite{wu2023visual}.


On the other hand, the latest abductive learning (ABL) framework  \cite{zhou2019abductive} unifies perception and reasoning in a mutually beneficial manner, which overcomes heavy-reasoning light-perception deficiencies of \cite{de2008probabilistic}, and solves heavy-perception light-reasoning flaws of \cite{getoor2007introduction}. Substantial advances have been made in areas such as theft judicial sentencing \cite{ouyang2022sentence}, stroke evaluation in table tennis \cite{wang2021tac}, neuro-symbolic learning tasks \cite{cai2021abductive,dai2020abductive,huang2021fast}. Nevertheless, it is insufficient for ABL to extract semantic information of individual reasoning rules expressed by logical clauses and their collaborative relationships, which are reflected in the following aspects: 1) construction of the knowledge base requires expert knowledge and complex transformations related to logical clauses; 2) representing mutual coupling information among logical rules is burdensome, particularly when attempting to match newly added reasoning rules with existing ones; 3) the joint optimization of perception and reasoning modules requires a huge amount of computation.

The aim of this paper is to provide a scheme called ChatABL, which combines the strength of LLMs and ABL framework to address the PLR problem in a more user-friendly and understandable manner. In this scheme, images are fed into perception module, i.e.,  various neural networks. Since the generated high-dimensional tensor cannot be understood by LLMs, we convert them into incomplete logical facts rendered in natural language format, conveying the partial depiction of the external environment by perception model and the accuracy is subject to verification. Then, the textual representations serve as corrective information for LLMs, which synthesize contextual cues, reasoning rules, and sample databases to realize constraint and rectification functionality. And the original perception data and the corrected results are used to supervise materials to update perception model. In turn, the optimized perception model provides the necessary logical facts for LLMs in natural language format, which can be used to strengthen LLMs. Finally, as a preliminary effort, the variable-length handwritten equation deciphering task, the well-known "holy grail" problem of artificial intelligence field \cite{dai2019bridging}, is regarded as a testbed in this work to testify the validity of the proposed ChatABL. Experimental results and comprehensive analysis with other mainstream methods validate the superiority of this method showing remarkable reasoning ability in solving complex reasoning problems.

The proposed method uses LLM’s robust logical reasoning capabilities to boost perception models, which allows us to extract the information of large unlabeled datasets. Another advantage of ChatABL is that the comprehensive information among extended rules and original domain knowledge can be leveraged to provide interactive reasoning explanations in natural language format. Overall, the main contributions of our work are summarized as follows:
\begin{itemize}
  \item [1)] 
  The ChatABL firstly explores a solution for bridging perception, language understanding, and reasoning ability via natural language interaction with LLM, providing insights for future research. The proposed method can complete complex reasoning tasks under the condition of small-sample data and incomplete knowledge.
  \item [2)]
  A novel knowledge-constrained self-feedback optimization strategy is designed utilizing penalty-based dynamics prompt. The proposed method utilizes the penalty-based dynamics prompt to execute trial-and-error and reasoning steps iteratively for refining logical facts by introducing self-feedback mechanism.  
  \item [3)]
  We have preliminarily rectified the previous erroneous conception by leveraging LLM to solve the handwritten equation deciphering task: LLM-based reasoning is a fallacy, and LLM chiefly serves as a user interface.
\end{itemize}


\section{Related Works}
\subsection{Perception and logical reasoning systems, and abductive learning}
It is suggested that human perception system and logical reasoning system are independent and mutually reinforcing \cite{aksyuk2023consciousness,ferilli2023gear,sloman1996empirical,xu2023mind}. Further, Marianna B. argues that these two systems work together to form our understanding of the world around us, with each system reinforcing and supporting the other \cite{ganapini2022combining}.

As one of the holy grail problems in AI, combining machine learning and logical reasoning has drawn much attention \cite{janiszewski2022abductive,lin2022exploit,magnani2022abductive,sapir2022machine}. Most existing methods try to combine the two different systems by making one side to subsume the other \cite{de2015probabilistic,getoor2007introduction,zadeh1965fuzzy}. Another typical approach is to use deep neural networks or other differentiable functional calculations to approximate symbolic calculi \cite{garcez2007abductive,towell1994knowledge}. However, most of them still require human-defined symbols as input \cite{russell2015unifying}. Few of them can make full-featured logical reasoning, and they usually require large amounts of training data. 

Recently, an abductive learning(ABL) method proposed by Zhou et.al  \cite{zhou2019abductive} targeted at unifying these two AI paradigms in a mutually beneficial way. The ABL method has been applied to a handwritten equation decipherment (HED) task, an abstract expression of the Mayan calendar decoding \cite{houston2001decipherment}, designed by Zhou and colleagues for demonstration, and the agent models can recognise numbers and resolve unknown mathematical operations simultaneously from images of simple hand-written equations, and can be generalised to longer equations and adapted to different tasks.





\begin{figure}[htbp]%
\setcaptionwidth{4.1in}
\centering
\includegraphics[width=0.6\textwidth,height=0.55\textwidth]{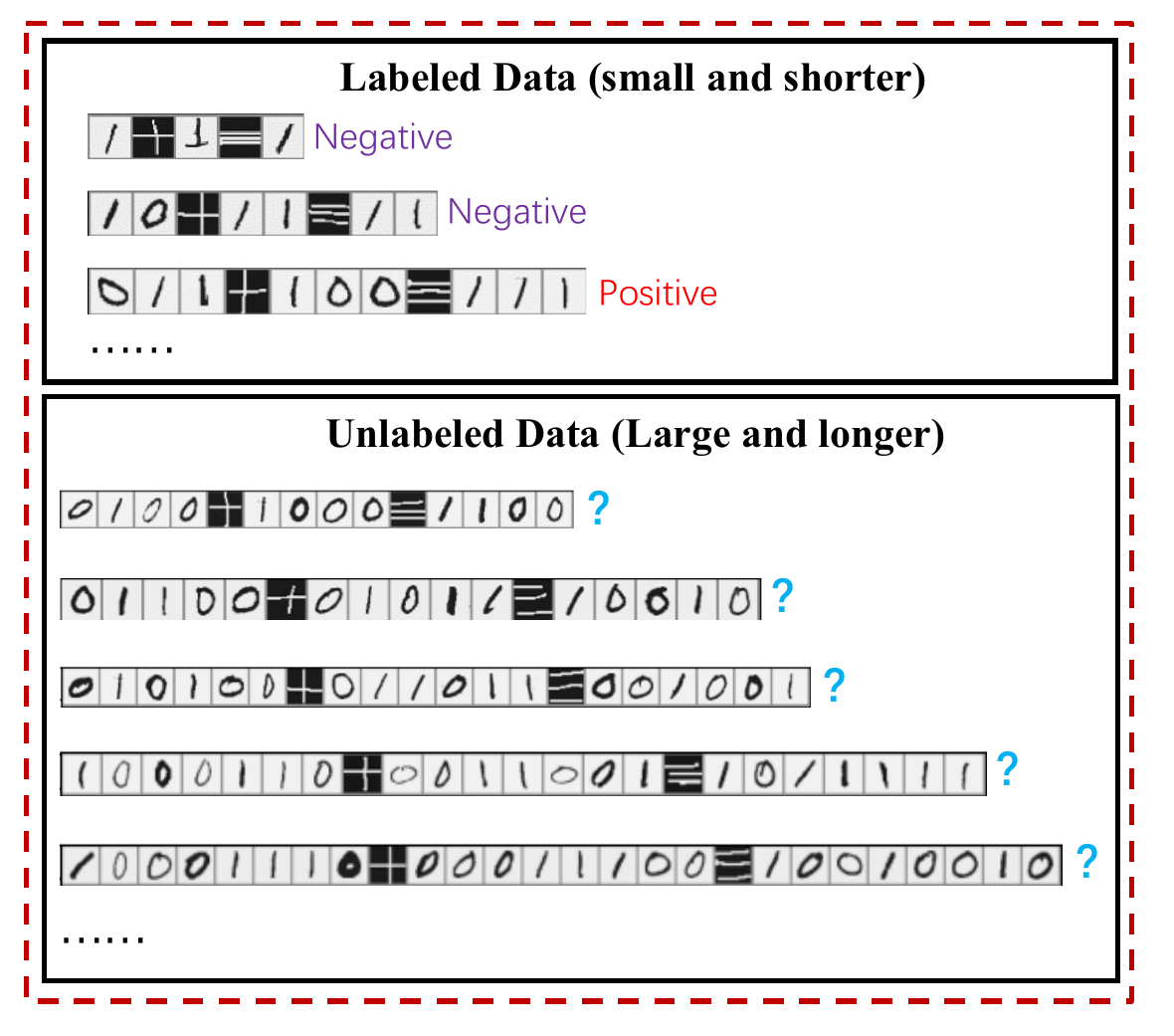}
\caption{Illustration of the handwritten equation decipherment, whose task is to recognise symbols and discover the unknown mathematical operation behind the handwritten equations from labeled images of small and shorter  equations, and can be generalised to unlabeled images of large and longer equations.
}\label{fig1}
\end{figure}

The handwritten equation decipherment task mentioned (\textbf{Figure \ref{fig1}}) is indeed a highly condensed and classic case that reflects the ideas of connecting perception and reasoning in intelligent behavior \cite{cai2021abductive,dai2018tunneling}. The equations are constructed from images of symbols (“0”, “1”, “+” and “=”), and they are generated with unknown operation rules, each example is associated with a label that indicates whether the equation is correct. A machine is tasked with learning from a training set of labelled equations, and the trained model is expected to predict unseen equations correctly.  

As described above, the small sample size and variable length equations make the resolution of this task challenging. The ABL framework tries to address these challenges by connecting machine learning with an abductive logical reasoning module and bridging them with consistency optimisation \cite{dai2019bridging}. The framework mainly consists of three parts: machine learning, logical abduction and optimisation. In the HED task, a CNN-based machine model \cite{hossain2018recognition} was used to generate pseudo-labels from image pixels. Logical abduction was achieved via an abductive logic program with Prolog computer language, and the RACOS optimization tool \cite{yu2016derivative}, using derivative-free optimisation algorithm, was deployed to solve the consistency optimization problem between pseudo-labels and background knowledge, which was designed as a logic program \cite{cai2021abductive}, involved only equation structure and recursive bit-wise operation definitions. Concretely, the background knowledge about equation structures is a set of Definite Clause Grammar (DCG) rules \cite{abramson1984definite,pereira1980definite}, which recursively defined digits as sequences of "0" and "1", and each equation had the format of X+Y=Z, allowing for varying lengths of X, Y, and Z. The logic program operating in a digit-by-digit manner reversed X+Y using the last digit as the starting point \cite{dai2018tunneling}.

The fact can be found that there are shortcomings in dealing with complex reasoning tasks by ABL method. The logic program could be too complex to interpret since it always involves complicated transformation with all kinds of logical predicate, hampering ABL's performance in generating accurate hypotheses and explanations. Therefore, it is necessary to find a new method to simplify the design and update of logical program.



\subsection{Large Language Models and ChatGPT}
As one of the most influential LLMs today, ChatGPT provides a user-friendly human-machine interaction platform and API that brings the powerful capabilities of large language models to the public and has been rapidly integrated into various fields of application such as education, healthcare, and others to perform general natural language processing tasks \cite{liu2023summary,dai2023chataug,wang2023chatcad,ma2023impressiongpt}, including text classification, data augmentation, arithmetic reasoning, sentiment analysis, question and answering, summarization, and more.

Since the last century, researchers have been exploring the ability of machines to process natural language. Early methods relied on human induction of knowledge from data and teaching machines to perform tasks. Later, with the advent of supervised learning, machines were able to learn automatically from annotated data. However, the amount of annotated data is limited compared to the vast amount of unannotated data, making it difficult for machines to learn universal knowledge. This changed with the emergence of pre-trained language models (PLMs), typically based on the Transformer architecture, and incorporating self-supervised learning. These models first learn universal knowledge from massive amounts of unannotated data, and then fine-tune on a small amount of annotated data, such as BERT  \cite{devlin2018bert}, Google T5  \cite{raffel2020exploring}, BART  \cite{lewis2019bart}, and OpenAI GPT  \cite{radford2018improving} series, with a parameter size of up to 1 billion.

After that, large language models (LLMs) with parameter sizes of over 100 billion have emerged, and they are still based on the Transformer architecture, but with vastly increased training data, parameter sizes, and model sizes. Researchers were surprised to find that as parameter size increases, the models exhibit new capabilities and continuous improvement in accuracy, which has led to the development of more LLMs, such as FLAN \cite{longpre2023flan}, GPT-3 \cite{brown2020language}, OPT \cite{zhang2022opt}, Bloom \cite{scao2022bloom}, PaLM \cite{chowdhery2022palm}, all of which have parameter sizes over 100 billion, with PaLM having 540 billion parameters.

Recent research \cite{cai2021chestxraybert,ma2023impressiongpt,wang2023chatcad} shows that ChatGPT have excellent high-level reasoning abilities, which can help researchers simplify experimental design and handle various non-symbolic problems. For example, Microsoft Research \cite{bubeck2023sparks} found that GPT-4 can solve a series of mathematical reasoning problems from elementary school to university level. In this paper, our goal is to use ChatGPT to replace the reasoning module in the reverse translation learning process to determine if the output of the perception module aligns with the rules of the knowledge base. 

\subsection{Reasoning via LLM}
The general capabilities and wide applicability of LLMs incentivized researchers to explore high-level reasoning abilities. In fact, reasoning is considered to be one of the emergent abilities when language models scale up in size \cite{bubeck2023sparks,wei2022emergent}.

The original GPT-3 model \cite{brown2020language} has demonstrated much potential in common sense reasoning through in-context learning, which lays the foundation for future reasoning research with LLMs. In addition, Wei et al. \cite{wei2022chain} hypothesized and verified that when given carefully prepared prompts sequentially (chain-of-thoughts), LLMs can perform significantly better in arithmetic reasoning, deductive reasoning and common sense reasoning through decomposition of multi-step problems. 

Recently, Wu et al. \cite{wu2023exploring}compared the deductive reasoning abilities of large language models. This study investigates ChatGPT and GPT-4's performance in the specialized domain of radiology, using a natural language inference task and comparing them to fine-tuned models. Results reveal that GPT-4 outperforms ChatGPT, and smaller fine-tuned models (e.g., BERT) require significant data to achieve GPT-4 level performance. The findings imply that creating a generic reasoning model based on LLMs for diverse tasks across various domains is viable and practical for clinical applications.

Another work by Ma et al. \cite{ma2023impressiongpt} explored LLM's ability to comprehend radiology reports through an innovative dynamic prompting paradigm. The impression section of radiology reports is crucial for communication between radiologists and other physicians. However, it is costly to produce valid impressions from radiology findings and this process calls for automation. The authors utilize LLMs' in-context learning ability by creating dynamic contexts with domain-specific, individualized sample findings-impression pairs. This approach enables the model to acquire contextual knowledge from semantically similar examples in existing data. They also develop an iterative optimization algorithm for automatic evaluation and prompt composition to further refine the model. ImpressionGPT achieves state-of-the-art performance on MIMIC-CXR and OpenI datasets without additional training data or LLM fine-tuning, presenting a localization paradigm for LLMs applicable across various domains with specific language processing requirements.

However, there is a scarcity of research on applying LLMs to abductive reasoning. One study by Jung et al. \cite{jung2022maieutic} improved LLMs' ability to make logical explanations through \textit{maieutic prompting}, which elicits abductive explanations to a problem through by recursively presenting the model with its own generated output as the inquiry. However, this study is limited to the scope of QA-style reasoning. 

This work is among the first efforts to present a comprehensive abductive learning framework that not only avoid the difficult process of symbolic rule formulation but also enhance the model's interpretability. We believe that the combination of these two factors will demonstrate the potential of ChatGPT in reasoning problems.

\section{Proposed Method}
\subsection{Problem Setting}
In the ChatABL framework for the HED task, the given input is defined as 
$\textbf {Input}=\left\{\boldsymbol{X}_l, \boldsymbol{X}_u,  \boldsymbol{K} \boldsymbol{B}_{\boldsymbol{\theta}}\right\}$,
where tensor $\boldsymbol{X}_l$ denotes labeled data, tensor
$\boldsymbol{X}_u$ denotes unlabeled data, the data size of
$\boldsymbol{X}_u$ is much larger than that of $\boldsymbol{X}_l$, and $\boldsymbol{KB}_{\theta}$
 denotes knowledge base, which is used to
constrain incomplete logical facts generated by perception module.
Concretely,
$\boldsymbol{X}_l=\left\{\left(\boldsymbol{x}_{l1}, \boldsymbol{y}_{l1}\right),\left(\boldsymbol{x}_{l2}, \boldsymbol{y}_{l2}\right), \ldots,\left(\boldsymbol{x}_{li}, \boldsymbol{y}_{l i}\right)\right\}$,
where variable ${x}_{li}$ represents shorter variable-length
handwritten equations, and $\boldsymbol{y}_{l i}$ implies the corresponding
labels. And the assignment of \(\mathbf{X}_{l}\) is to learn a mapping
from $\boldsymbol{x}$ to $\boldsymbol{y}$. 
$\boldsymbol{X}_u=\left\{\boldsymbol{x}_{u 1}, \boldsymbol{x}_{u 2}, \ldots, \boldsymbol{x}_{u j} \mid i \ll j\right\}$,
where $\boldsymbol{x}_{u j}$ denotes the unlabeled handwritten equations
with longer length, are utilized to boost representative
capability of the above mapping. $\boldsymbol{KB}_{\theta}$ consists of a
series of domain rules in natural language format with learnable
objective $\boldsymbol{\theta}$ in LLMs, which integrate fewer labeled
data $\boldsymbol{X}_l$ and a large amount of unlabeled data
$\boldsymbol{X}_u$ to optimize perceptual model $\boldsymbol{f}$ and mine
unknown rules $\boldsymbol{\theta}$ for handwritten equation with the
constraint of knowledge base using LLMs. The ChatABL algorithm yields
the corresponding pseudo-labels to the unlabeled data by the classifier
optimized by a small amount of labeled signals, and the produced labels
may be incorrect due to the small number of training samples, which is
difficult to guarantee good performance. Therefore, the ChatABL modifies
the pseudo-labels and learns the reasoning rules of the knowledge base
at the same time by LLMs, so that the consistency of them is maximized
under the constraint of the knowledge base. Formally, the problem
definition can be summarized as an optimization problem of searching
\(\mathbf{Output}\) under a given \(\mathbf{Input}\):

\begin{equation}
\begin{aligned}\label{eq1}
\min _{\boldsymbol{f}, \boldsymbol{\theta}}\qquad  
 \operatorname{Loss}_{l a b e l}\left(\boldsymbol{y}_{l i}, \boldsymbol{f}_{l i}\right)+\operatorname{Loss}_{\text {unlabel }}\left(\boldsymbol{\delta}\left(\boldsymbol{y}_{u j}^{\prime}\right), \boldsymbol{f}_{u j}\right) \\
 \text { s.t. } \underset{\boldsymbol{\delta}}{\operatorname{argmax}} \text { constraint }\left(\boldsymbol{\delta}\left(\boldsymbol{y}_{u j}^{\prime}\right), \boldsymbol{f}_{l i}, \boldsymbol{K} \boldsymbol{B}_\theta\right)
 \end{aligned}
 \end{equation}

where \(\mathbf{y}_{uj}^{'}\) is the pseudo-label corresponding to the
\(j^{th}\) unlabeled instance, which is generated by the perceptual
module. $\boldsymbol{\delta}\left(\cdot\right)$  indicates an implicit heuristic
function learned by LLM, which aims to revise pseudo-labels by logical
reasoning process. In addition to correcting inconsistent pseudo-labels,
this goal also helps the knowledge base to find unknown rules
$\boldsymbol{\theta}$. It can be seen from \textbf{Eq.\ref{eq1}}  that the major
challenge is how to mine the effective information of more massive and
complex unlabeled image data under the
$\boldsymbol{KB}_{\theta}$ constraints, and react to the iterative update of itself and reason the underlying mathematical laws behind logical events. Noted that LLMs can perform reasoning tasks based on their understanding of language without requiring optimization. It can be founded from \textbf{Eq.\ref{eq1}} that the introducing of LLM to the aforementioned problem will result in transformation of the optimization goal for ABL method, from the joint optimization problem of discrete variables and continuous variables to the optimization problem of continuous variables only for the perception model. The proposed issues in the \textbf{Introduction} part of this paper may be potentially addressed from this perspective. Further research details will be elaborated in the following sections.



\subsection{Overall Framework}
ChatABL takes the images of variable length equations as input, extracts and embeds element features from the images, and outputs the judgment results of these equations and their underlying law. The key problems during constructing ChatABL are as follows.

\textbf{1) KP1: How to make use of image information in limited labeled variable-length handwritten equations?} Labeled large and longer handwritten equations are limited due to their high demand for proficient domain knowledge, which can be seen from the prototype problem (the deciphering process of the Mayan calendar \cite{houston2001decipherment}) that it is difficult for the most outstanding expert to directly give a definite answer or even solve it in most cases when facing the strange number. Therefore, the rational use of a small amount of labeled data poses a huge challenge to the generalization ability of agent systems.

\textbf{2)	KP2: How can we learn to leverage existing clues to tackle higher-level reasoning in complex reasoning tasks?} In the case of unknown arithmetic operations, it is challenging for many individuals, particularly those unfamiliar with the domain, to find the law of unlabeled handwritten equations with limited rules and small and short labeled handwritten images. Therefore, how to effectively mine the existing conditions and learn the rules embedded within them to accomplish more challenging reasoning is an intriguing topic.

\textbf{3)	KP3: How to combine the perceptual information from images with domain knowledge rules in understandable manner?} It is difficult to inject symbolic knowledge into the optimization of numerical values in machine learning models \cite{wang2021tac}. The major obstacle is that the current methods face difficulties in representing and synthesizing rules. Moreover, their corresponding reasoning processes fall short in terms of interpretability, rendering them challenging to express in natural language.

\begin{figure}[htbp]%
\setcaptionwidth{4.1in}
\centering
\includegraphics[width=0.87\textwidth]{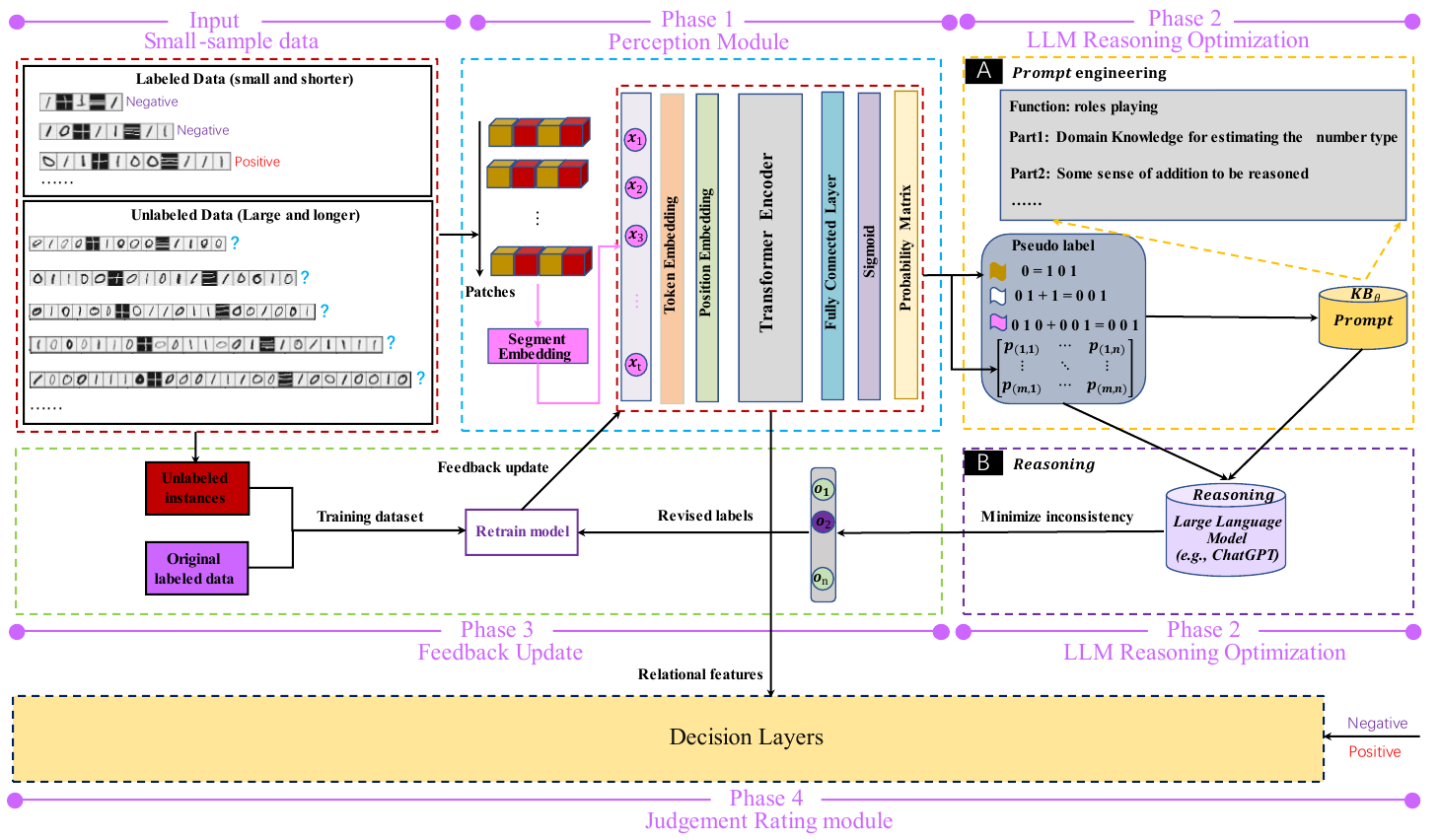}
\caption{Schematic representation of the ChatABL framework for HED task, which leverages LLM's logical reasoning ability to correct the understanding of the images of the equations from the perception module, thereby optimizing the recognition of symbols and the mining ability of the mathematical operation behind them.}\label{fig2}
\end{figure}

Considering the aforementioned issues, we constructed ChatABL into four phases including Perception Module, LLM Reasoning Optimization, Feedback Update and Judgement Rating module (\textbf{Figure \ref{eq2}}), which leverage ABL combined with LLM to extract and embed the variable length equations attributes using limited labeled data. \textbf{Phase 1} involves using a recognition component as perception module to embed the equations. Then we introduce a logical reasoning component (\textbf{KP1}) based on domain knowledge rules described in natural language in \textbf{Phase 2}, which leverages reasoning ability of large language models to strengthen pseudo-labels of unlabeled data (\textbf{KP2}). In \textbf{Phase 3}, the raw perception data and the revised labels are then used as supervision materials to iteratively improve the performance of the attribute recognition component in a feedback update manner (\textbf{KP3}). Finally, a classifier is trained in \textbf{Phase 4}, which learns  evaluation in image-based methods to obtain quantified evaluation results. We judge the veracity of handwritten equations into two levels as training labels and train the model to classify the level of unlabeled equations based on the embedding vectors in \textbf{Phase 1}.


The model trained in \textbf{Phase 3} is the same as the perception model introduced in \textbf{Phase 1}, and cross-entropy loss is utilized for its training \cite{zhang2018generalized}. Additionally, a conventional fully-connected layer is used for the decision layer, which has been elaborated elsewhere \cite{dai2018tunneling}. This paper first provides a brief introduction on the perception model and then focuses on the LLM reasoning optimization.

\subsection{Perception Model}
The perception model serves as the backbone for recognizing image inputs. The primary goal of this perception model is to extract visual features from the data. Specifically, the extracted features will be used to construct logical facts and mathematical expressions, which are used for subsequent processing. In this study, we employ the Vision Transformer (ViT) \cite{dosovitskiy2020image} as the perception model to process the handwritten inputs. It is noteworthy that our framework is compatible with any vision model.

The ViT, originally adapted from the Transformer architecture in the domain of natural language processing, has demonstrated exceptional performance in various computer vision tasks, often surpassing traditional convolutional neural networks (CNNs). By treating an image as a sequence of tokens, the ViT effectively captures both local and global contextual information through its self-attention mechanism. Leveraging this powerful architecture allows us to effectively discern complex patterns and intricate spatial relationships present in handwritten mathematical expressions. 


To maintain the independence of mathematical expressions in the handwritten input (the \textbf{Input} part in \textbf{Figure \ref{fig2}}), we split them into individuals $\boldsymbol{X}_u=\left\{\boldsymbol{x}_{u 1}, \boldsymbol{x}_{u 2}, \ldots, \boldsymbol{x}_{u n}, \ldots, \boldsymbol{x}_{u j} \right\}$ using segment embedding. Meanwhile, we reshape $\setlength{\lineskiplimit}{2.5pt} \boldsymbol{x}_{u n}\in\boldsymbol{R}^{\boldsymbol{H}\times\boldsymbol{W}\times\boldsymbol{C}}$ into a sequence of flattened patches $\boldsymbol{X}_{upn}\in\boldsymbol{R}^{\boldsymbol{N}\times\boldsymbol{P}^{2}\times\boldsymbol{C}}$ to serve as input for the ViT. $\boldsymbol{N}=\frac{\boldsymbol{HW}}{\boldsymbol{P}^{2}}$ and $\boldsymbol{P}$ denote the number of patches and the resolution of image patches, respectively. Every mathematical expression $\boldsymbol{X}_{upn}\in\boldsymbol{R}^{\boldsymbol{N}\times\boldsymbol{P}^{2}\times\boldsymbol{C}}$ is made up of several patches representing binary digits ("0" and "1") or mathematical operators. The ViT generate pseudo-labels for subsequent processing as follows:
\begin{equation}\label{eq2}  \boldsymbol{z}_{0}=\lbrack\boldsymbol{x}_{cls};\boldsymbol{x}_{up1}\boldsymbol{L};\boldsymbol{x}_{up2}\boldsymbol{L};\ldots;\boldsymbol{x}_{upn}\boldsymbol{L};\ldots;\boldsymbol{x}_{upN}\boldsymbol{L}\rbrack + \boldsymbol{L}_{pos}
\end{equation}

\vspace{-10pt}

\begin{footnotesize}
\begin{equation}\label{eq3}
\begin{split}
     \boldsymbol{z}_{m}&=\boldsymbol{MLP}(\boldsymbol{LN}(\boldsymbol{MSA}(\boldsymbol{LN}(\boldsymbol{z}_{m-1}))+\boldsymbol{z}_{m-1})) \\
    & +\boldsymbol{LN}(\boldsymbol{MSA}(\boldsymbol{LN}(\boldsymbol{z}_{m-1}))+\boldsymbol{z}_{m-1})
\end{split}
\end{equation}
\end{footnotesize}

\begin{footnotesize}
\begin{equation}\label{eq4}
    \boldsymbol{y}^{'}=\boldsymbol{SIG}(\boldsymbol{FCL}(\boldsymbol{z}_{M}))
\end{equation}
\end{footnotesize}


\vspace{5pt}

where $\boldsymbol{z}_{0}$ is the latent vector mapped by a trainable linear projection from a sequence of flattened patches, consisting of token embedding and position embedding $\boldsymbol{L}_{pos}$ (\textbf{Eq.\ref{eq2}}). $\lbrack\boldsymbol{z}_{m},\boldsymbol{m}=1,2,\ldots,\boldsymbol{M}\rbrack$ represents the results of the transformer encoder at different layers, which includes multiheaded self-attention (MSA), multilayer perceptron (MLP), and layernorm (LN) (\textbf{Eq.\ref{eq3}}). $\boldsymbol{y}^{'}$ generates the pseudo-label with a probability matrix based on the output from $\boldsymbol{z}_{M}$ (\textbf{Eq.\ref{eq4}}). $\boldsymbol{SIG}$ and $\boldsymbol{FCL}$ denote the fully connected layer and sigmoid, respectively.


\subsection{LLM Reasoning Optimization}

In this work, we employ penalty-based dynamic prompt and rules-constrained self-feedback optimization strategy to enhance the adaptation of LLM for complex reasoning tasks. The main concept behind our approach is inspired by the human process of solving complex reasoning tasks, with "solving math problems" (SMP) being an illustrative example. The process involves interpreting known conditions and transforming them into several advantageous conditions for efficient problem-solving. Usually, mathematical conditions involved in complex problems are not typically encountered previously, or the specific forms utilized may vary from those previously encountered. In such cases, we rely on reasoning abilities to perform multiple iterations of trial and reasoning, under given rules and examples. If our reasoning results conform with the rules, we pass the test; if not, we would receive reminders indicating contradictions or deviations from the rules, which guides further reasoning and decision-making. In essence, the capacity of advanced human reasoning is characterized by iterative reasoning under rule constraints and reminders. This abstract representation of how humans solve complex reasoning tasks inspires us to design the relevant prompts and optimization strategies to enhance LLMs' powerful logical reasoning abilities. Overall, our method requires a small number of examples, facilitating LLM’s robust logical reasoning ability with limited rules. Further details will be presented in the subsequent sections.


\subsubsection{Penalty-based Dynamic Prompt}
Before introducing the Penalty-based Dynamic Prompt, we first explicate the input structure of LLM, which is accessible via a string to the API and divided into three components: System, User, and Assistant. The System message is typically invoked at the outset to demarcate the task and constrain the Assistant's behavior, while the User message serves to provide direction and constitutes the user's input. The model output is generated by the Assistant component, which serves as the basis for our dynamic prompt and iterative optimization framework.

Prior studies have used fixed-form prompts for straightforward tasks that could be easily generalized. However, these prompts lack the necessary prior knowledge for complex tasks and domain-specific datasets, resulting in low performance \cite{ma2023impressiongpt}. Thus, we propose a hypothesis that constructing dynamic prompts in a reward-punishment manner from relevant domain-specific corpora can enhance the model's comprehension and perception.

\begin{figure*}[htbp]%
\setlength{\belowcaptionskip}{-2mm}
\setcaptionwidth{4.1in}
\centering
\includegraphics[width=0.87\textwidth]{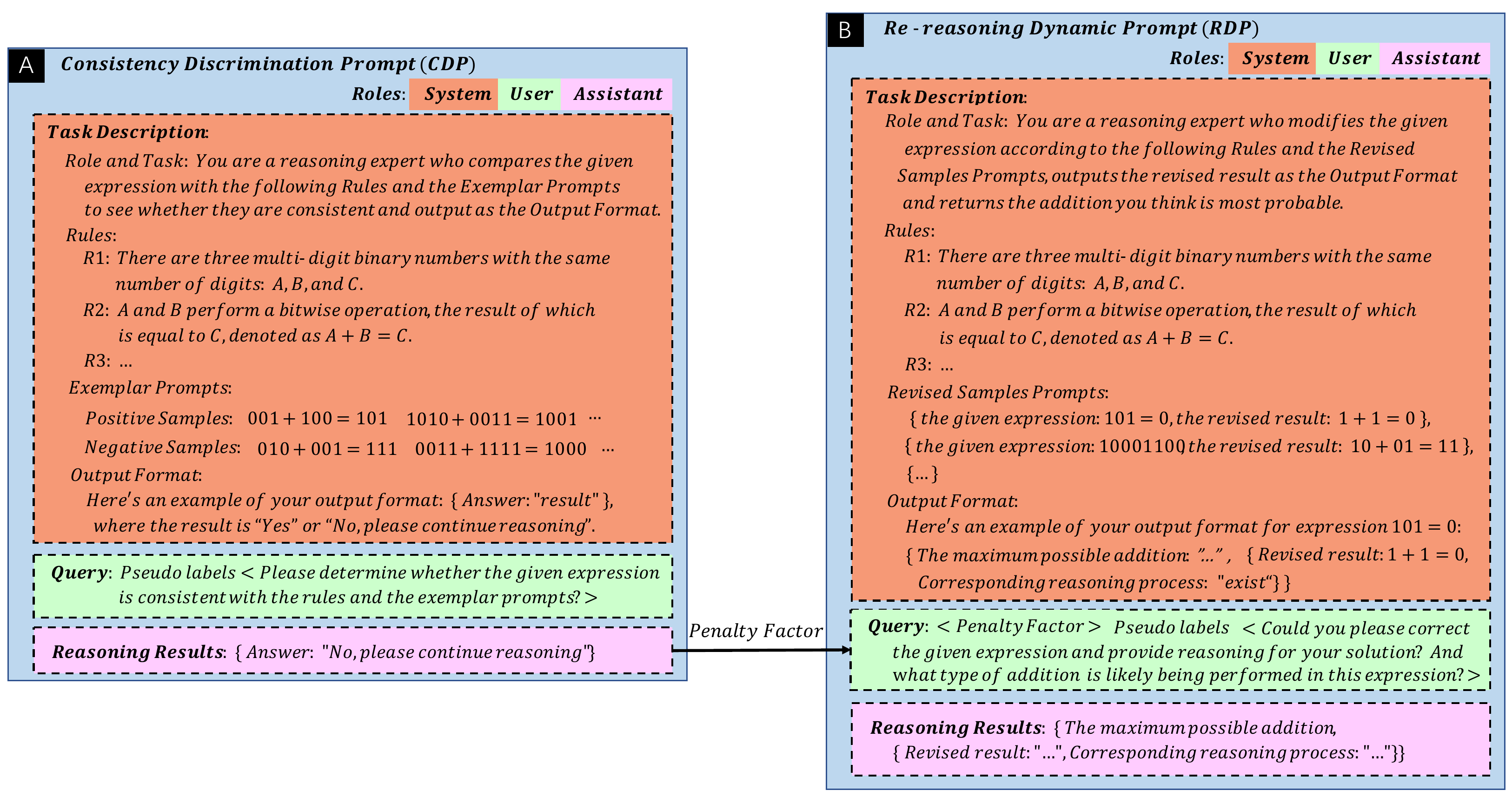}
\caption{Details of penalty-based dynamic prompt, which consists of consistency discrimination prompt (CDP) and re-reasoning dynamic prompt (RDP). The CDP reminds whether to perform the reasoning process RDP by the penalty factor. They contains task description, query and reasoning results components.}\label{fig3}
\end{figure*}

Inspired by SMP process, when solution result is inconsistent with existing rules, reminders are given to indicate the contradiction. These reminders correspond to the prompt of LLMs for adaptive adjustment. Therefore, the prompt can be updated  by using this kind of feedback information as a guide for LLMs. Based on the premise, two prompts are designed in this paper: consistency discrimination prompt (CDP) and re-reasoning dynamic prompt (RDP), as shown in \textbf{Figure \ref{fig3}(A)(B)}. The primary function of the CDP is to determine whether the pseudo-labels 

\noindent obtained by the perceptual model are consistent with the given rules and examples. If inconsistency is detected, it serves as a reminder for the RDP (corresponding to the SMP process) to adjust dynamically in a punishment way. If consistency is detected, there is no need for the RDP. Both prompts are composed of task description, query and reasoning results. In the CDP, we first assigned the role of a reasoning expert to LLM for task description, after which we presented the reasoning task accompanied by a set of rules, exemplar prompts, and output format. The significance of this configuration lies in its ability to define the precise nature of the task undertaken by the model, serving as the fundamental framework for the entire prompt. After that, we generate the prefix query sentence: "Please determine whether the given expression is consistent with the rules base and the exemplar prompts ?", which is used to further specify consistency discrimination task. At the end of CDP, the reasoning result is integrated into the RDP query as a penalty factor. In the RDP, the task description format is similar to that of CDP, but with minor variations as the RDP accomplishes reasoning and correction tasks. Specifically, the rules applied in this task are consistent with those of CDP. The reasoning task is designed to correct a given expression. \textbf{Figure \ref{fig3}(B)} illustrates the exemplar prompts, and the output format comprises the corrected expression and its corresponding reasoning process. After that, we generate the query sentence: "Could you please correct the given expression and provide reasoning for your solution ? And what type of addition operation is likely being performed in this expression?", which is used to further specify re-reasoning task. Note that the penalty factor from CDP needs to be added to the query before proceeding. At the end of the dynamic prompt, the results produced in this manner utilize the inconsistency between the given rules and pseudo-labels to strengthen the LLM's reasoning direction and enhance its performance.


\subsubsection{Knowledge-Constrained Self-Feedback Optimization}

\begin{figure*}[htbp]%
\setcaptionwidth{4.1in}
\centering
\includegraphics[width=0.87\textwidth,height=0.87\textwidth]{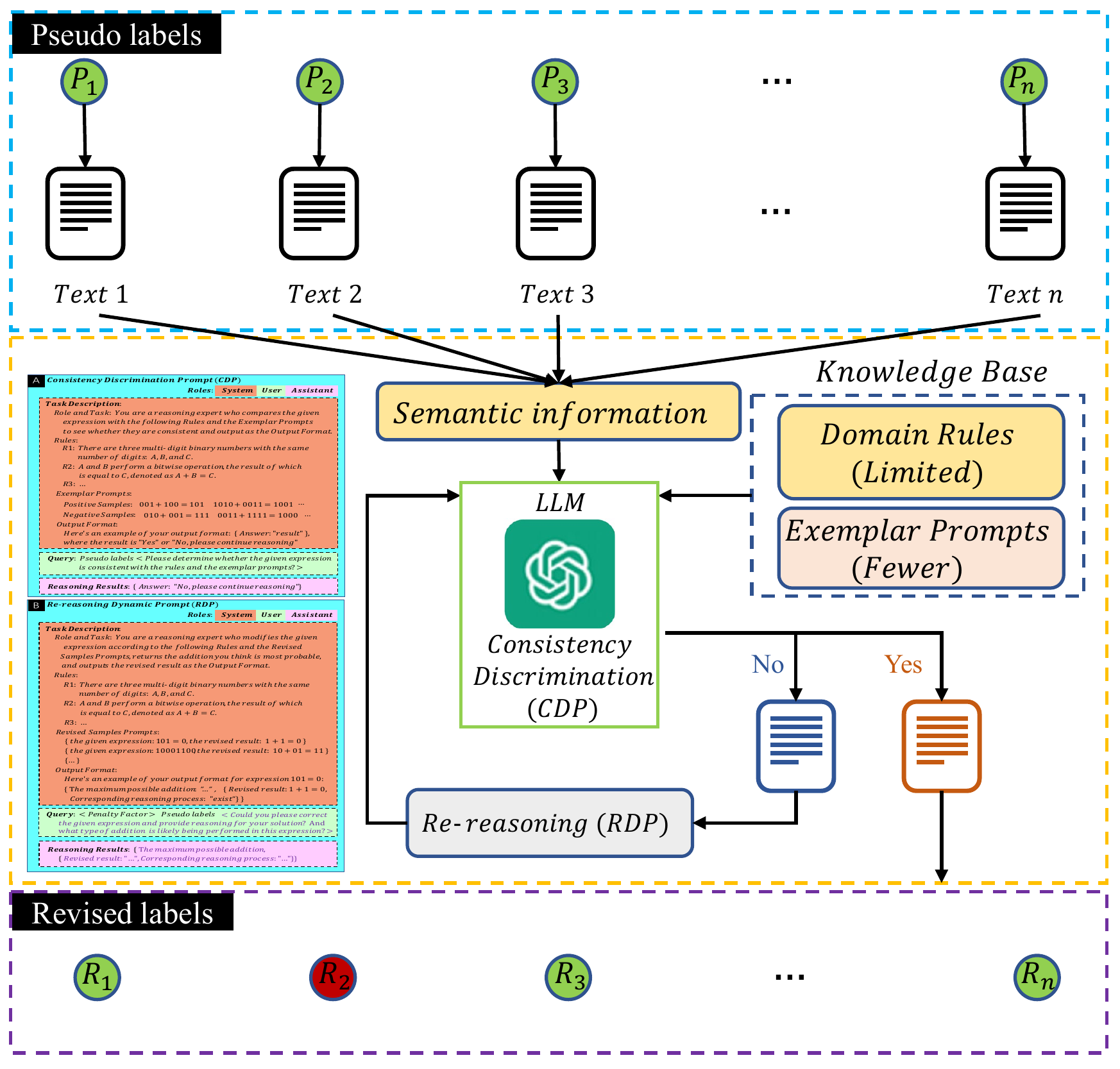}
\caption{Details of knowledge-constrained self-feedback optimization strategy, which utilizes LLM with the penalty-based dynamics prompt to execute trial-and-error and reasoning steps iteratively for refining pseudo-labels by the consistency between abductive information and knowledge base.}\label{fig4}
\end{figure*}

In \textbf{section 3.4.1}, it can be seen that the design of the penalty-based dynamic prompt highlights a one-time effectiveness for LLM, which differs from humans' multiple trial-and-error reasoning process when solving complex mathematical problems. This limitation may lead to uncertainty regarding whether the LLM's response aligns with our anticipated outcomes. This paper presents a novel knowledge-constrained self-feedback optimization strategy that leverages incomplete but certain knowledge, including existing rules and fewer examples, as discriminative criterion for reasoning incomplete facts, specifically pseudo-labels that require further validation. The proposed method utilizes the penalty-based dynamics prompt to execute trial-and-error and reasoning steps iteratively for refining pseudo-labels by introducing self-feedback mechanism. The epoch of iterations is dynamically determined based on consistency results between incomplete facts and knowledge, eliminating the requirement for laborious and time-consuming manual intervention.

In detail, the proposed method has been elaborately illustrated in \textbf{Figure \ref{fig4}}. Firstly, the pseudo-labels generated by the perception model are transformed into corresponding text labels (e.g., combinations of symbols "0", "1", "+", and "="), followed by semantic interpretation of the text information. To determine whether the correction process is required, we utilize the consistency between the semantic interpretation and the knowledge base (the same as the rules and exemplar prompts fields described in \textbf{section 3.4.1}) , which relies on the CDP, described in \textbf{Figure \ref{fig3}(A)}. However, in the limited labeled data, the perception model's ability to generate consistent pseudo-labels with respect to the rules and exemplar prompts for new instance is often inadequate. In such case, we use the corresponding reasoning results: “No, please continue reasoning”, and it is produced by the CDP to identify inconsistencies, which serves as the query part of the RDP for re-reasoning task, as illustrated in \textbf{Figure \ref{fig3}(B)}. Subsequently, the reasoning results generated by the LLM with RDP are repeatedly fed back into the CDP to validate consistency until the revised labels are in alignment with the knowledge base, which are eventually outputted as the revised labels. 


The advantage of this knowledge-constrained self-feedback approach lies in the fact that the LLM conducts abducive reasoning on incomplete facts under the constraint of incomplete knowledge. This enables LLM to provide a reliable guarantee for the perception model in terms of knowledge constraints within the existing domain knowledge system, while also iteratively updating its own abstraction description of logical facts. As a result, the reasoning capability of LLM is significantly strengthened, which enhances its effectiveness in complex logical reasoning situations with incomplete knowledge.


\section{Experimental Study}
\subsection{Experiment Setup}

We construct a sequence of raw images of digits and operators to the hand writing equations based benchmark handwritten character datasets \cite{lecun1998gradient,thoma2017hasyv2}, as shown in \textbf{Figure \ref{fig2}}. We furnish ChatABL according to \textbf{Figure 3} with background domain knowledge in natural language format regarding arithmetic structural rules \cite{dai2018tunneling}. It is noteworthy that these rules do not contain information about the type of computation performed within the equations; instead, ChatABL has to derive that information from the available data. An illustrative example is depicted in \textbf{Figure \ref{fig3}}. We evaluate the learning performance of ChatABL by benchmarking it against Meet [12], CNN-BiLSTM \cite{dai2018tunneling,hochreiter1997long}, Transformer(TF) \cite{vaswani2017attention}, and ABL \cite{dai2019bridging}, which involves pure perception methods (the former three), perception and reasoning methods (the last one). These state-of-the-art models are capable of addressing sequential input tasks. The training set utilized for all methods comprises equations ranging in length from 5 to 10, with each length containing 500 randomly-sampled equations. During the testing phase, all methods are tasked with predicting 5000 equations, ranging in length from 5 to 26, with 500 examples for each length. The same configuration is employed for prompt, reasoning optimization, and feedback updating. In our experiments, we adopt Accuracy, Precision, Recall, F1, and AUC \cite{fu2020multisensor} as evaluation metrics in our experiments, which essentially cover most of the metrics in the field.

\subsection{Comparison on Small Samples Performance}
\vspace{-13pt}

\begin{table*}[htbp]%
\centering
\setcaptionwidth{4.1in}
\caption{Performance comparison of the ChatABL method to state-of-the-art methods as the proportion change of training dataset in terms of metrics ($\%$) for handwritten equation recognition task.}\label{tab1}
\resizebox{0.84\textwidth}{!}
{
\begin{tabular}{l c c c c c}
\noalign{\smallskip}
\toprule[2pt]
\noalign{\smallskip}
\makebox[0.2\textwidth][l]{Method/Metrics} &  \makebox[0.1\textwidth][c]{Accuracy} &
\makebox[0.1\textwidth][c]{Precision} &
\makebox[0.1\textwidth][c]{Recall} &
\makebox[0.1\textwidth][c]{F1} &
\makebox[0.1\textwidth][c]{AUC}\\
\noalign{\smallskip}
\hline
\noalign{\smallskip}
MEET-20 &51.53 &60.86 & 50.68 &53.71 &50.41\\
\noalign{\smallskip}
CNN-BiLSTM-20 &54.71 & 66.94 &52.84 &59.06 &53.19\\
\noalign{\smallskip}
\textbf{ChatABL-20} &\textbf{68.92} &\textbf{78.01} &\textbf{66.90} &\textbf{69.43} &\textbf{68.44}\\
\noalign{\smallskip}
ABL-20 &91.36 &98.28 &87.34 &91.93 &87.85\\
\noalign{\smallskip}
\hline
\noalign{\smallskip}
MEET-50 &54.70 &69.91 &53.13 &60.38 &53.63\\
\noalign{\smallskip}
CNN-BiLSTM-50 &56.44 &64.75 &54.55 &59.22 &56.56\\
\noalign{\smallskip}
\textbf{ChatABL-50} &\textbf{70.31} &\textbf{79.23} &\textbf{69.56} &\textbf{71.32} &\textbf{70.44}\\
\noalign{\smallskip}
ABL-50 &93.18 &98.95 &88.81 &93.61 &93.05\\
\noalign{\smallskip}
\hline
\noalign{\smallskip}
TF-80 &63.18 &86.76 &58.49 &69.87 &64.63\\
\noalign{\smallskip}
MEET-80 &64.47 &90.45 &59.27 &71.62 &65.18\\
\noalign{\smallskip}
\bottomrule[2pt]
\noalign{\smallskip}
\end{tabular}
}
\vspace{-6pt}
\end{table*}

\textbf{Table \ref{tab1}} presents the experimental results of performance metrics for different comparative methods as the proportion changes in the training set, which consists of three subsets in the case of 20$\%$, 50$\%$ and 80$\%$ labeled data as the training set. Compared to the ABL method based on first-order logical formula reasoning, the performance of ChatABL is lower than that of ABL, because the ABL method revises the error iteratively between images about handwriting equations and knowledge base by optimizing joint consistency on continuous and discrete variables. In contrast, the method proposed in this paper does not need to involve the optimization of complex discrete variables. The corresponding rules are described by natural language description, and then the perception module is modified by using the reasoning ability of LLM, which also achieves good performance. This indicates that LLM has an advantage in unifying perception and understanding reasoning tasks through natural language interaction.

On the other hand, by comparing the results of the pure perception method and ChatABL when the labeling rate is low (the first two sections in \textbf{Table \ref{tab1}}), the gains of using both unlabeled data and knowledge rules for ChatABL are much higher. Further comparison on the third section shows that the ChatABL performance with 20$\%$ and 50$\%$ training set can even be superior the methods (TF,MEET) with 80$\%$ training set. The results verified that logic consistency can be very useful for providing a surrogate supervised signal through LLM reasoning process for revising the pseudo-labels from perception module. In short, LLMs based on pre-trained language models work better than previous studies, which allows the model to adequately learn the prior knowledge of the complex reasoning domain.

\subsection{Evaluation on Perceptual Model}

In order to validate the performance effect of the perception module in the overall ChatABL framework on the HED tasks, we test the AlexNet \cite{krizhevsky2017imagenet}, GoogleNet \cite{Szegedy_2015_CVPR}, ResNet \cite{DBLP:journals/corr/HeZRS15}, and ViT \cite{dosovitskiy2020image} models as the perception component of the ChatABL framework, respectively, because they are the most advanced benchmark models in the development of convolutional neural networks on solving tasks from image input. In particular, the ViT surpasses traditional CNNs in various computer vision tasks, because it treats images as token sequences that capture both local and global contextual information through self-attention mechanism. Additionally, we use the same metrics as the \textbf{4.2 section} for comparison.
\vspace{-10pt}
\begin{table*}[htbp]%
\centering
\setcaptionwidth{4.1in}
\caption{The influence of different perception modules and GPTs on ChatABL as the 20$\%$ proportion of training dataset in term of metrics ($\%$).}\label{tab2}
\resizebox{0.84\textwidth}{!}
{
\begin{tabular}{l c c c c c}
\noalign{\smallskip}
\toprule[2pt]
\noalign{\smallskip}
\makebox[0.2\textwidth][l]{Method/Metrics} &  \makebox[0.1\textwidth][c]{Accuracy} &
\makebox[0.1\textwidth][c]{Precision} &
\makebox[0.1\textwidth][c]{Recall} &
\makebox[0.1\textwidth][c]{F1} &
\makebox[0.1\textwidth][c]{AUC}\\
\noalign{\smallskip}
\hline
\noalign{\smallskip}
AlexNet	&64.17	&72.23	&58.98	&61.95	&63.86\\
\noalign{\smallskip}
GoogleNet	&66.54	&76.58	&62.19	&67.09	&65.96\\
\noalign{\smallskip}
ResNet	&66.44	&78.98	&59.35	&66.26	&64.51\\
\noalign{\smallskip}
ViT	&68.92	&78.01	&66.90	&69.43	&68.44\\
\noalign{\smallskip}
\hline
\noalign{\smallskip}
GPT-3.5-Turbo &68.92 &78.01 &66.90 &69.43 &68.44\\
\noalign{\smallskip}
GPT-4 & 70.01 & 79.51  & 67.82 & 70.63 & 69.85 \\
\noalign{\smallskip}
\bottomrule[2pt]
\noalign{\smallskip}
\end{tabular}
}
\vspace{-5pt}
\end{table*}

As illustrated in the first part of \textbf{Table \ref{tab2}}, the overall performance impact of ChatABL is affected by different perception modules. In general, the performance improves progressively as better perception methods, only except that ChatABL with ResNet as the perception module presents slightly worse performance compared to that with GoogleNet. And the approach of utilizing ViT as the perception model for ChatABL ultimately achieved the best performance among the comparative methods, with an accuracy rate approaching 70$\%$. This represents an advancement in the context of HED tasks. Based on the results, it can be founded that an improved perception accuracy indeed impacted the performance of equation classification by ChatABL, which is similar to the effect of human sensory system on the whole central nervous system.

\subsection{The Impact of Different GPTs}

In this section, we primarily compared the impact of the latest large language models on ChatABL. Since small-scale language models with limited parameters cannot provide meaningful reasoning processes \cite{wang2023chatcad}, they were excluded from this experiment. We selected GPT-3.5-Turbo and GPT-4 as the reasoning component in the second phase of the ChatABL method. The corresponding parameter quantities of the two models are approximately 175 billion and 1 trillion, respectively \cite{brown2020language}. Their parameter magnitudes differ by almost three orders of magnitude, thus comparing the impact of the two models on the overall performance is sufficient.

The lower part of \textbf{Table \ref{tab2}} reports the performance metrics. It can be seen that the ChatABL reasoning module's transition from GPT-3.5-Turbo to GPT-4 leads to an overall increase in all metrics, particularly with in greatest gain in accuracy from 68.92$\%$ to 70.01$\%$. Taken together with the findings in \textbf{section 4.2}, these results demonstrate significant progress in studying the generalization ability of modern intelligent systems with small-sample data, and in synthesizing perception and reasoning via natural language as a medium. Overall, the reasoning capability of language models is proportional to their size, highlighting the critical factor of the logistic reasoning capability of LLMs.

\section{Discussion and Conclusion}
This paper first explore a novel framework, ChatABL, which strives to achieve integration of perception, language understanding, and reasoning capabilities in user-friendly manner. To achieve this goal, we propose the incorporation of large language models into abductive learning, and design the prompt akin to human problem-solving, to enhance the effectiveness of ChatABL in complex mathematical reasoning task. And we have fundamentally rectified the previous erroneous conception: LLM-based reasoning is a fallacy and LLM chiefly serves as user interface. 

However, the proposed method still has limitations to be solved. Compared to humans, the performance of large language models is heavily dependent on the context and design of the prompt. And there is significant room for improvement in prompt engineering in the future. Additionally, in our study, we acknowledge that there are areas where this paper lacks rigor. We have conducted a qualitative analysis of prompt design, but we have not done a quantitative analysis. And the number of tokens in LLM is limited, which may affect the performance of the model. Once the latest model, GPT5.0, is released in the future, we will conduct further in-depth investigation. Furthermore, we envision expanding ChatABL to encompass multimodal data perception and integrated reasoning with diverse domain knowledge, providing novel directions for further research.






\bibliographystyle{splncs04}
\bibliography{mybib} 
\end{sloppypar}
\end{document}